\documentclass{llncs}

\usepackage{times}
\usepackage{graphicx}
\usepackage{marvosym}
\usepackage{amsfonts,amsmath,amssymb}
\usepackage[utf8]{inputenc}
\usepackage{tikz}
\usetikzlibrary{arrows,snakes,backgrounds}
\usepackage{float,flafter}
\usepackage{url}

\usepackage[ruled]{algorithm2e} 

\SetAlFnt{\scriptsize}
\SetAlCapFnt{\scriptsize}
\SetAlCapNameFnt{\scriptsize}
\SetAlCapHSkip{0pt}
\IncMargin{-\parindent}

\SetKwFor{Loop}{loop}{}{end}
\SetKwFor{uLoop}{loop}{}{}
\SetKwFor{Let}{let}{}{end}
\SetKwFor{uLet}{let}{}{}
\SetKwFor{myWhile}{while}{do}{}
\SetKwFor{myForEach}{foreach}{do}{}
\SetKwFor{myFor}{for}{do}{}

\SetKwFunction{kernelEGinitialize}{kernel\_EG\_initialize}
\SetKwFunction{kernelEGliftA}{kernel\_EG\_lift1}
\SetKwFunction{kernelEGliftB}{kernel\_EG\_lift2}
\SetKwFunction{removenulls}{remove\_nulls}
\SetKwFunction{clear}{clear}
\SetKwFunction{kernelEGliftAshfl}{kernel\_EG\_lift1\_shfl}

\SetKwFunction{InitialPropagation}{InitialPropagation}
\SetKwFunction{PropagateAndCheck}{PropagateAndCheck}
\SetKwFunction{Learning}{Learning}
\SetKwFunction{Decision}{Decision}
\SetKwFunction{Backjump}{Backjump}
\SetKwFunction{CompleteAssignment}{CompleteAssignment}

\SetKwFunction{mkdlbitmap}{mk\_dl\_bitmap}

\SetKwFunction{syncthreads}{\_\_syncthreads}
\SetKwFunction{reset}{reset}
\SetKwFunction{fwdlearning}{fwd\_learning}
\SetKwFunction{leftmostsetbit}{leftmost\_set\_bit}
\SetKwFunction{getithdecision}{get\_ith\_decision}
\newcommand{\dependencies}{{\mathit{Deps}}}
\newcommand{\insieme}[2]{\{{#1}  \mid  {#2}\}}
\newenvironment{des}{
\begin{list}
{$\bullet$}
{\topsep = 1 mm
\labelwidth = 2 mm
\labelsep = 2 mm
 \parsep = 0.1 mm
\itemsep = \parskip
\leftmargin = 6 mm}
}{\end{list}}

\date{}
\title{GPU-based parallelism for ASP-solving{\,}\thanks{Research supported by 
INdAM-GNCS-19 project,
by Univ.\,of~Perugia (projects ``ricerca-di-base-2016--18'', YASMIN, CLTP, and RACRA), and Univ.\,of Udine PRID~ENCASE.}}

\author{
Agostino Dovier\inst{1}
\and
Andrea Formisano\inst{2}
\and
Flavio Vella\inst{3}}
\institute{Universit\`a  di Udine  \and
Universit\`a  di Perugia \and
Libera Universit\`a di Bolzano}

\begin{document}

\maketitle

\begin{abstract}
Answer Set Programming (ASP) has become, the paradigm of choice in the field of logic programming
and non-monotonic reasoning. Thanks to the availability of
efficient solvers, ASP has been successfully
employed in a large number of application domains.
The term GPU-computing indicates a recent programming paradigm aimed at enabling the use of 
 modern parallel Graphical Processing
Units (GPUs) for general purpose computing.
In this paper we describe an approach to ASP-solving that exploits GPU parallelism.
The design of a GPU-based solver poses various challenges due to the peculiarities of GPUs' software and hardware architectures
and to the intrinsic nature of the satisfiability problem.
\begin{keywords}
ASP solvers, ASP computation, SIMT parallelism, GPU computing
\end{keywords}
\end{abstract}

\section*{Introduction}

Answer  Set Programming (ASP) is an expressive and purely declarative framework 
developed in the last decades in the Logic Programming and Knowledge Representation communities.
Thanks to its  extensively studied mathematical foundations and the continuous improvement of
efficient and competitive solvers, ASP has become the paradigm of choice in many fields of AI.
It has been fruitfully employed in many areas, 
such as knowledge representation and reasoning, planning, bioinformatics, multi-agent systems, 
data integration, language processing, declarative problem solving, semantic web, robotics, 
to mention a few among many~\cite{ErdemGL16,Falkner2018}.

The clear and highly declarative nature of ASP enables excellent opportunities for the introduction 
of parallelism and concurrency in implementations of ASP-solvers.
Steps have been made in the last decade  toward the parallelization of the basic components 
of Logic Programming systems. Implementations of solvers 
exploiting multicore architectures, distributed systems, or portfolio approaches,
have been proposed~\cite{DFP18}.
In this direction, a recent new stream of research concerns the design and development 
of parallel ASP systems that can take advantage of the massive degree  of parallelism 
offered by modern Graphical Processing Units (GPUs). 

GPUs are multicore devices designed to operate with very large number of lightweight threads,
executing in a rigid synchronous manner. 
They present a significantly complex memory organization. 
To take full advantage of GPUs'
computational power, one has to adhere to specific programming directives, in order to
proficiently distribute the workload among the computing units and achieve the highest 
throughput in memory accesses.
This makes the model of parallelization used on  GPUs deeply different from those employed in 
more ``conventional'' parallel architectures.
For these reasons, existing parallel solutions are not directly applicable in the context of GPUs.

This paper illustrates the design and implementation of a conflict-driven ASP-solver
that is capable of exploiting the \emph{Single-Instruction Multiple-Thread}
parallelism offered by GPUs. 
As we will see, the overall structure of  the GPU-based solver
is reminiscent of the  conventional structure of sequential 
conflict-driven  ASP solvers  (such as, for example, the state-of-the-art solver {\sc clasp}~\cite{Gebser2012Practice}).
However, substantial differences lay in both the implemented algorithms and in the adopted programming model.
Moreover, we avoid two hardly parallelizable and intrinsically sequential components usually present in existing solvers.
On the one hand, we exploit \emph{ASP computations} to avoid the introduction of loop formulas and the need of
performing \emph{unfounded set checks}~\cite{Gebser2012Practice}. 
On the other hand, we adopt a parallel conflict analysis procedure as an alternative to the sequential resolution-based technique used in {\sc clasp}. 

The paper is organized as follows. Sect.~\ref{sect:preliminary} recalls basic notions on ASP, GPU-computing, and the CUDA framework.
The approach to ASP solving based on \emph{conflict-driven nogood learning} is described in Sect.~\ref{sect:CDASP}.
Sect.~\ref{sect:IA} illustrates the difficulties  inherent in parallelizing irregular applications, such as ASP, on
GPUs.
The software architecture of the CUDA-based ASP-solver {\sc yasmin} is outlined in Sect.~\ref{sect:yasmin}. 
In particular, the new parallel learning procedure is presented in Sect.~\ref{sect:Learning} and evaluated in Sect.~\ref{sect:runs}.

\section{Preliminaries}\label{sect:preliminary}
We briefly recall the basic notions on ASP needed in the rest of the paper 
(for a detailed treatment see~\cite{gelfond2007,Gebser2012Practice} and the references therein).
Similarly, we also recall few needed notions on CUDA parallelism~\cite{cuda19,CUDAZONE}.
\subsubsection*{Answer Set Programming.~}
An ASP program $\Pi$ is a set of ASP rules of the form:
\\\centerline{$
p_0 \leftarrow p_1, \dots, p_m, \textit{not } p_{m+1}, \dots, \textit{not } p_n
$}
where $n \geq 0$ and each $p_i$ is an atom. If $n=0$, the rule is a \emph{fact}.
If $p_0$ is missing, the rule is a \emph{constraint}. Notice that such a  constraint can be rewritten as a
headed rule of the form $q \leftarrow p_1, \dots, p_m, \textit{not } p_{m+1}, \dots, \textit{not } p_n, \textit{not } q$,
where $q$ is a fresh atom.
Hence, constraints do not increase the expressive power of ASP.

A rule including variables is simply seen as a shorthand for the set of its ground instances.
Without loss of generality, in what follows we consider the case of ground programs only.
(Hence, each $p_i$ is a propositional atom.)

Given a rule $r$, 
$p_0$ is referred to as the \emph{head} of
the rule  ($head(r)$), while the set 
 $\{ p_1, \dots, p_m, \textit{not } p_{m+1}, \dots, \textit{not } p_n \}$
is referred to as the \emph{body} of $r$ ($body(r)$). Moreover, we put
$body^{+}(r) = \{p_1, \dots, p_m\}$, 
 $body^{-}(r) = \{p_{m+1}, \dots, p_n\}$, $\varphi^{+}(r)= p_1\wedge \dots \wedge p_m$ and
$\varphi^{-}(r)=\neg p_{m+1}\wedge  \dots\wedge  \neg p_n$.
We will 
denote the set of all atoms in $\Pi$ by~$atoms(\Pi)$ and the set of all rules
defining the atom~$p$ by
$rules(p) = \insieme{r}{head(r)=p}$.
The \emph{completion} $\Pi_{cc}$ of a program $\Pi$ is defined as the formula:
\\\centerline{$
\Pi_{cc} = 
\bigwedge_{p \in atoms(\Pi)}
\Big(
p ~\leftrightarrow~ 
\bigvee_{r \in rules(p)}
\big(  \varphi^{+}(r) \wedge \varphi^{-}(r) \big)   
\Big)
$.}

Semantics of ASP programs is expressed in terms of \emph{answer sets}.
An \emph{interpretation} is a set $M$ of atoms; $p\in M$ (resp. $p \not\in M$) denotes
that $p$ is true (resp. false). An interpretation is a \emph{model} of
a rule $r$ if $head(r)\in M$, or $body^{+}(r)\setminus M \neq \emptyset$, 
or $body^{-}(r) \cap M \neq \emptyset$. $M$ is a model of a program $\Pi$ if it is
a model of each rule in~$\Pi$.
$M$ is an \emph{answer set} of $\Pi$ if it is the subset-minimal model of the \emph{reduct program}~$\Pi^M$.

An important connection
exists between the answer sets of $\Pi$ and the minimal models of~$\Pi_{cc}$.
In fact, any answer set of $\Pi$ 
is a minimal model of~$\Pi_{cc}$.
The converse is not true, but it can be shown~\cite{LinZhao03} that the answer sets of
$\Pi$ are the minimal models of~$\Pi_{cc}$
satisfying the \emph{loop formulas} of~$\Pi$. 
The number of loop formulas can be, in general, exponential in the size of~$\Pi$.
Hence, modern ASP solvers adopt some form of \emph{lazy} approach to generate loop formulas only ``when needed''.
We refer the reader to~\cite{LinZhao03,Gebser2012Practice} for the details;
in what follows we will describe an alternative approach to answer set computation that avoids the generation of loop formulas.
The new approach 
exploits \emph{ASP computations} to avoid the introduction of loop formulas and the need of performing \emph{unfounded set checks}~\cite{Gebser2012Practice}
during the search of answer sets.

\smallskip

The notion of ASP computations originates from a computation-based characterization of answer sets~\cite{ASPcomputation}
based on an incremental construction process, where at each step choices determine which
rules are actually applied to extend the partial answer set. 
More specifically, for a program $\Pi$ let $T_\Pi$ be the immediate consequence operator of $\Pi$.
Namely, if $I$ is an interpretation, then
\\\centerline{$
T_\Pi(I)  =  \big\{ {head(r)}   \,\mid\,   r \in \Pi \, \wedge body^{+}(r)\subseteq I \wedge  body^{+}(r)\cap I = \emptyset \big\}
$.}
An \emph{ASP Computation} for $\Pi$ is a sequence of 
interpretations $I_0, I_1, I_2, \dots$ (where $I_0$ can be any set of atoms that are logical consequences of~$\Pi$)
satisfying these conditions:
\begin{des}
\item[\mbox{\sc persistence of beliefs:~}]
	$I_i \subseteq I_{i+1}$ for all $i\geq 0$ 
\item[\mbox{\sc convergence:~}]
	$I_\infty = \bigcup_{i=0}^{\infty} I_i$ is such that $T_\Pi(I_\infty) = I_\infty$;
\item[\mbox{\sc revision:~}]
	$I_{i+1} \subseteq T_\Pi(I_i)$ for all $i \geq 0$;
\item[\mbox{\sc persistence of reason:~}]
	if $p\in I_{i+1} \setminus I_i $ then there is $r\in rules(p)$ such that $I_j$ is a model of $body(r)$ for each
    $j \geq i$.
\end{des}
Following~\cite{ASPcomputation}, 
an interpretation $I$ is an answer set of~$\Pi$
if and only if there exists an ASP computation such that $I = \bigcup_{i=0}^{\infty} I_i$.

\subsubsection*{GPU-computing and the CUDA framework.}
\emph{Graphical Processing Units (GPUs)} are massively  parallel devices,
originally developed to support efficient computer graphics and fast image processing.
The use of such multicore systems has become pervasive
in general-purpose applications that are not directly related to computer graphics,
but demand massive computational power.
The term \emph{GPU-computing} indicates the use of the modern GPUs for such general-purpose computing.
NVIDIA is one of the pioneering manufacturers in promoting GPU-computing, especially
through the support to its \emph{Computing Unified Device Architecture (CUDA)}~\cite{CUDAZONE}.
A GPU  contains hundreds or thousands of identical computing units (\emph{cores}) and
provides access to both on-chip memory (used for registers and shared
memory) and off-chip memory (used for cache and global memory).
Cores are grouped in 
a collection of \emph{Streaming MultiProcessors (SMs);} in turn, each SM contains a
fixed number of computing cores.

The underlying conceptual model for parallelism
is \emph{Single-Instruction Multiple-Thread (SIMT),} where the same instruction is
executed by different threads that run on cores, while data and operands may differ from thread to thread.
A logical view of computations is introduced by CUDA, in order to define
abstract parallel work and to schedule it among different hardware configurations.
A typical CUDA program is a C/C++ program
that includes parts meant for execution on the CPU (referred to as the \emph{host}) and
parts meant for parallel execution on the GPU (referred to as the \emph{device}).

The CUDA API supports interaction, synchronization, and communication between host and device.
Each device computation is described as a collection of concurrent threads, each executing the same device function
(called a \emph{kernel}, in CUDA terminology).
These threads are hierarchically organized in \emph{blocks} of threads and \emph{grids} of blocks.
The host program contains all the instructions needed 
to initialize the data in  the GPU,  specify the number of grids/blocks/threads, and manage the kernels.
Each thread in a block executes an instance of the kernel, and has a thread ID within its block.
A grid is a 3D array of 
blocks that execute the same kernel, read data input from the global memory, and
write results to the global memory.
When a CUDA program on the host  launches a kernel, the blocks of the grid are 
scheduled to the SMs with available execution capacity.
The threads in the same block can share data, using  high-throughput on-chip shared memory, while
threads belonging to different blocks can only share data through the global memory. Thus,
 the block size allows the programmer to define the granularity of threads cooperation.

It should be noticed that the most efficient access pattern to be adopted by threads in reading/storing data 
depends on the kind of memory.
We briefly mention here two possibilities (see~\cite{cuda19} for a comprehensive description).
Shared memory is organized in \emph{bank}s. 
In case threads of the same block accesses locations in the same bank, a \emph{bank conflict} occurs and the 
accesses are serialized. To avoid bank conflicts, \emph{strided} access pattern has to be adopted. 
On the contrary, concerning global memory, to reach the highest throughput, \emph{coalesced} accesses have to be executed.
Intuitively, this can be achieved if consecutive threads access contiguous global memory locations.

Threads of each block are grouped in \emph{warp}s of 32 threads each.
The threads of the same warp share the fetch of the instruction code to be executed. Hence, the maximum efficiency is achieved when all
32 threads execute the same instruction (possibly, on different data).
Whenever two (or more) groups of threads belonging to the
same warp fetch/execute different instructions, \emph{thread divergence} occurs.
In this case the execution of the different groups is serialized and the overall performance decreases.

A simple CUDA application presents the following basic components:\footnote{Notice that,
for the sake of simplicity, we are ignoring many aspects of CUDA programming and advanced
techniques such as  dynamic parallelism, cooperative groups, multi-device programming, etc.
We refer the reader to~\cite{cuda19} for a detailed treatment.}
\begin{des}
\item[{\sc memory allocation and data transfer.}]
Before being processed by kernels, data must be copied to the global memory of the device.
The CUDA API supports memory allocation and data transfer to/from the host.

\item[\sc kernels definition.]
Kernels are defined as standard C functions; the annotation used to communicate to the CUDA compiler that a function should be treated as kernel has the form:~
{\tt\small\_\_global\_\_ void kernelName (Formal Arguments)}.

\item[\sc kernels execution.]
A kernel can be launched from the host
program using:\\
\centerline{\tt\small kernelName <<< GridDim, TPB >>> (Actual Arguments)}
 where 
{\tt\small GridDim} describes the number of blocks of the grid and  {\tt\small TPB} specifies the number of threads in each block.

\item[\sc data retrieval.]
After the execution of the kernel, the host  retrieves the results
with a transfer operation from global memory  to host memory.
\end{des}

\section{Conflict-driven ASP-Solving}\label{sect:CDASP}

Conflict-driven nogood learning (CDNL) is one of the techniques successfully used by ASP-solvers,
such as the {\sc clingo} system~\cite{Gebser2012Practice}.
The first attempt in exploiting GPU parallelism for conflict-driven
ASP solving has been made in~\cite{DFPV15_ICLP,DFPV16}.
The approach adopts a conventional architecture of an ASP solver which starts by translating the completion $\Pi_{cc}$
of a given ground program~$\Pi$ into a collection of \emph{nogoods} (see below).
Then, the search for the answer sets of $\Pi$ is performed by exploring a search 
space composed of all interpretations for the atoms in $\Pi$, organized as a binary tree.
Branches of the tree correspond to (partial) assignments of truth values to program atoms (i.e., partial interpretations). 
The computation of an answer set proceeds by alternating decision steps and propagation phases. 
Intuitively:
(1)~A decision consists in selecting an atom and assigning it a truth value.
	(This step is usually guided by powerful heuristics analogous to those developed for SAT~\cite{SAThandbook}.)
(2)~Propagation extends the current partial assignment by adding all consequences of the decision. 
The process repeats until a model is found (if any).
It may be the case that inconsistent truth values are propagated for the same atom after $i$ decisions (i.e., while visiting a node at depth~$i$
in the tree-shaped search space). 
In such cases a \emph{conflict} arises \emph{at decision level~$i$}
testifying that the current partial assignment cannot be extended to a model of the program.
Then, a \emph{conflict analysis} procedure is run to detect the reasons of the failure.
The analysis identifies which decisions should be undone in order to restore consistency of the assignment. It also
produces a new \emph{learned} nogood to be added
to the program at hand, so as to exclude repeating the same failing sequence of decisions, in the subsequent part of the computation.
Consequently, the program is extended with the learned nogood and the search backjumps to a previous (consistent) point in the search
space, at a decision level~$j<i$.
Whenever a conflict occurs at the top decision level ($i=1$), the computation ends because no (more) solutions exist.

Following~\cite{DFPV15_ICLP,DFPV16}, let us outline how CDNL can be combined with ASP computation in order to obtain a
solver that does not need to use loop formulas.
We describe  both assignments $A$ and  nogoods $\delta$ as
sets of \emph{signed atoms}---i.e., entities of the form $Tp$ or $Fp$,
denoting that $p\in atoms(\Pi)$ has been assigned {\tt true} or {\tt false}, respectively.
Plainly, assignment contains at most one element between $Tp$ and $Fp$ for each atom~$p$.
Given an assignment $A$, let $A^T = \insieme{p}{Tp \in A}$.
Note that $A^T$ is an interpretation for~$\Pi$.
A \emph{total} assignment $A$ is such that, for every atom $p$, $\{Tp,Fp\}\cap A\neq\emptyset$. 
Given a (possibly partial) assignment $A$ and a nogood $\delta$, we say that $\delta$ is
\emph{violated} if $\delta \subseteq A$.
In turn, $A$ is a \emph{solution} for a set of nogoods $\Delta$ if no $\delta \in \Delta$ is violated by~$A$.
Nogoods can be  used to perform deterministic propagation (\emph{unit propagation}) and extend an assignment.
Given
a nogood $\delta$ and a partial assignment~$A$ such that $\delta \setminus A = \{Fp\}$ (resp.,~$\delta \setminus A = \{Tp\}$),
then we can infer the need to add $Tp$ (resp.,~$Fp$) to~$A$ in order to avoid violation of~$\delta$.

Given a program $\Pi$, a set of {\it completion nogoods} $\Delta_{\Pi_{cc}}$ is derived from~$\Pi_{cc}$ as follows.
For each  rule $r\in\Pi$ and each atom $p \in atoms(\Pi)$, we introduce the formulas: 
\\\centerline{$
\begin{array}{rcl c rcl c rcl c rcl}
b_r & \leftrightarrow & t_r \wedge n_r &   ~~~~~&
t_r & \leftrightarrow & \varphi^{+}(r) &   ~~~~~&
n_r & \leftrightarrow &  \varphi^{-}(r)   &   ~~~~~&
p   & \leftrightarrow & \bigvee_{r\in rules(p)} b_r \\
\end{array}
$}
where $b_r, t_r, n_r$ are new atoms 
(if $rules(p)=\emptyset$, then the last formula reduces to $\neg p$).
The completion nogoods reflect the structure of the implications in these formulas:
\begin{des} 
\item from the first formula we have the nogoods: 
$\{Fb_r,Tt_r, Tn_r\},
\{Tb_r,Ft_r\}$, and $\{Tb_r, Fn_r\}$.

\item {F}rom the second and third formulas we have the nogoods:
$\{Tt_r, Fp\}$ for each $p\in body^{+}(r)$; 
$\{Tn_r, Tq\}$ for each $q\in body^{-}(r)$; 
$\{Ft_r\} \cup \insieme{Tp}{p \in body^{+}(r)}$; and
$\{Fn_r\} \cup \insieme{Fq}{q \in body^{-}(r)}$.

\item {F}rom the last formula we have the nogoods:
$\{Fp, Tb_r\}$ for each $r \in rules(p)$ and
$\{Tp\}\cup \insieme{Fb_r}{r \in rules(p) }$.
\end{des}
Moreover, for each constraint $\leftarrow p_1, \dots, p_m, \textit{not } p_{m+1}, \dots, \textit{not } p_n$ 
in $\Pi$ we  introduce a \emph{constraint nogood} of the form $\{Tp_1,\dots,Tp_m,Fp_{m+1},\dots,Fp_n\}$.
The set $\Delta_{\Pi_{cc}}$ is the set of all the nogoods so defined.

The basic CDNL procedure described earlier can be easily combined with the notion of ASP computation.
Indeed, it suffices to apply a specific heuristic during the selection steps to satisfy the four properties defined in Sect.~\ref{sect:preliminary}.
This can be achieved by assigning true value to a selected atom only if this atom is supported by a rule with true body.
More specifically, let $A$ be the current partial assignment, the selection step acts as follows. 
For each unassigned atom $p$ occurring as head of a rule in the original program, 
all nogoods reflecting the rule $b_r \leftarrow t_r, n_r$, such that $r \in rules(p)$ are analyzed to check
whether $Tt_r \in A$ and $Fn_r \notin A$ (i.e., the rule is \emph{applicable}~\cite{ASPcomputation}).
One of the rules $r$ that pass this test is selected.
Then, $Tb_r$ is added to $A$. In the subsequent propagation phase 
$Tp$ and $Fn_r$ are also added to~$A$ and 
$Fn_r$ imposes that all the atoms  of   $body^{-}(r)$ are set to false.
This, in particular,  ensures the \emph{persistence of beliefs} of the ASP computation.
(In the real implementation (see Sect.~\ref{sect:yasmin}) all applicable rules $r$, and their heads, 
are evaluated according to a heuristic weight and the rule $r$ with highest ranking is selected.)
It might be the case that no selection is possible because no unassigned atom $p$ exists such that there is an applicable~$r\in rules(p)$.
In this situation the computation ends by assigning false value to all unassigned heads in~$\Pi$. This completes the assignment,
which is validated by a final propagation step in order to check that no constraint nogoods are violated. In the positive case the assignment 
so obtained is an answer set of~$\Pi$.

\section{ASP as an irregular application}\label{sect:IA}

The design of GPU-based ASP-solvers poses various challenges due to the structure and intrinsic nature of the satisfiability problem.
The same holds for GPU-based approaches to SAT~\cite{CUDASAT2}.
As a matter of fact, 
the parallelization of SAT/ASP-solving shares many aspect with other applications of GPU-computing where problems/instances are
characterized by the presence of large, sparse, and unstructured data.
Parallel graph algorithms constitute significant examples, that, like SAT/ASP solving,
exhibit irregular and low-arithmetic intensity combined with  data-dependent control flow and memory access patterns.
Typically, in these contexts, large instances/graphs have to be modeled and represented using sparse data structures
(e.g., matrices in \emph{Compressed Sparse Row/Column} formats).
The parallelization of such algorithms struggle to achieve scalability due to lack of data locality, 
irregular access patterns, and unpredictable computation~\cite{lumsdaine2007challenges}. 
Although, in the case of some graph algorithms, several techniques have been established in order to improve performance on 
parallel architectures~\cite{hong2011efficient} and accelerators~\cite{8267334},
the different character of the algorithms used in SAT/ASP might prevent from obtaining comparable
impact on performance by directly applying the same techniques.  
This is because, first, the time-to-solution of a SAT/ASP problem is dominated by heuristic selection and learning procedures able
to cut the exponential search space. In several cases, smart heuristics might be most effective than advanced parallel solutions. 
Second, because of intrinsic data-dependencies, procedures like propagation or learning often require to access large parts of the data/graph, sequentially.
Similarly to what experienced in other complex graph-based problems~\cite{FGV17},  the kind of computation involved differs from that of
traversal-like algorithms (such as, Breadth-First Search) which process a subset of the graph in iterative/incremental manners and for which
advanced GPU-solutions exist.
Furthermore, aspect specific to the underlying architecture enters into play, such as
coalesced memory access and CUDA-thread balancing, which are major objectives in parallel algorithm design.
In this scenario, our GPU-based proposal to ASP solving also implements:
\begin{des}
 \item efficient parallel propagation able to maximize memory throughput and minimize thread divergence. 
 \item Fast parallel learning algorithm which avoids the bottleneck represented by the intrinsically sequential
	 resolution-like learning procedures commonly used in CDNL solvers.
 \item Specific thread-data mapping solutions able to regularize the access to data stored in global, local, and shared memories.
 \end{des}
In what follows we will describe how to achieve these requirements in the  GPU-based solver for ASP.

{\begin{algorithm}[tb]
\DontPrintSemicolon
\CommentSty{\color{blue}}
	\caption{\label{alg:yasmin}Host code of the ASP-solver {\sc yasmin} ~~~\hfill(simplified)}
{\bf procedure} {\sc yasmin}({$\Delta$: SetOfNogoods, $P$: GroundProgram})\\
	\nl$cdl \gets 1$ ;~ \reset($A$)\tcc*{set initial decision level and empty assignment}
	\nl$\InitialPropagation\texttt{<<<b,t>>>}(A, \Delta, \mathit{Viol})$\tcc*[f]{check input units satisfaction}\\
\nl\lIf{$\mathit{Viol}$}{{\textbf{return $\mbox{no-answer-set}$}}}
\nl\lElse{\uLoop{}
{ 
	\nl$\PropagateAndCheck(A, \Delta, cdl, \mathit{Viol})$\tcc*{update\,{$A$} and flag $Viol$}
	\nl\lIf(\tcc*[f]{Violation at first dec.level}){$\mathit{Viol} \land (cdl = 1)$}{\textbf{return {$\mbox{no-answer-set}$}}}
	\nl   \ElseIf(\tcc*[f]{Violation at level $cdl{>}1$}){$\mathit{Viol}$}{
	\nl $\Learning\texttt{<<<b,t>>>}(\Delta, A, cdl)$\tcc*{conflict analysis: update $\Delta$ and $cdl$}
	\nl $\Backjump\texttt{<<<b,t>>>}(A, cdl)$\tcc*{update $A$ and $cdl$}
	  }
	\nl   \uIf{($A$ is not total)}{
		\tcc{rank selectable literals and applicable rules. If possible, select $Lit$, 
		extend $A$, update $cdl$. Otherwise,~$Lit\gets\mathit{nil}$\,:}
		\nl$\Decision\texttt{<<<b,t>>>}(\Delta, A, \mathit{Lit})$\\
		\nl    \uIf(\tcc*[f]{no applicable rules}){$\mathit{Lit}=\mathit{nil}$}{
			\nl$\CompleteAssignment\texttt{<<<b,t>>>}(A)$\tcc*[f]{falsify unassigned atoms}}
		 }
	\nl\lElse{{\bf return} {$A^T \cap atom(P)$\tcc*[f]{stable model found}}}
}
}
\end{algorithm}
}

\section{The CUDA-based ASP-solver {\sc yasmin}}\label{sect:yasmin}

In this section, we present a solver that exploits ASP~com\-pu\-ta\-tion,
nogoods handling, and GPU parallelism.
The ground program $\Pi$, as produced by the grounder {\sc gringo}~\cite{Gebser2012Practice}, is read by the CPU. 
The CPU also computes the completion nogoods $\Delta_{\Pi_{cc}}$ and transfers them to the device. 
The rest of the computation is performed completely  on the GPU.
During this process, there only memory transfers between the host and device
involve control-flow flags (e.g., an ``exit'' flag, used to communicate
whether the computation is terminated) and the computed answer set (from the GPU to the~CPU).

As concerns representation and storing of data on the device, nogoods are represented using 
Compressed Sparse Row (CSR) format. The atoms of each nogood are stored contiguously and 
all nogoods are stored in consecutive locations
of an array allocated  in global memory.
An indexing array contains the offset of each nogood, to enable direct accesses to them.
(Such indexes are then used as identifiers for the corresponding nogoods.)
Nogoods are sorted in increasing order, depending on their length. 
Each atom in $\Pi$ is uniquely identified by an \emph{index}, say~$p$.
A array $A$ of integers is used to store in global memory the set of assigned atoms (with their truth values) in this manner:
\begin{des}
\item $A[p]= 0$  if and only if the atom $p$ is unassigned;  
\item $A[p]= i$, $i > 0$ (resp., $A[p]= -i$) means that atom $p$ has been assigned 
{\tt true} (resp., {\tt false})  at the decision level~$i$.
\end{des}

The basic structure of the {\sc yasmin} solver is shown in Alg.~\ref{alg:yasmin}.
We adopt the following notation: 
for each signed atom $p$, let $\overline{p}$ represent the same atom with opposite sign.
Moreover, let us refer to the stored set of nogoods simply by the variable~$\Delta$.
The variable $cdl$ (initialized in line~1) represents the current decision level.
As mentioned, $cdl$ acts as a counter that keeps track of the current number of decisions that have been made.

Since the set of input nogoods may include some unitary nogoods, a preliminary parallel computation partially initializes~$A$ accordingly (line~2).
It may be the case that inconsistent assignments occur in this phase. In such case a flag $Viol$ is set, the given program $\Pi$ in declared unsatisfiable
(line~3) and the computation ends.
Notice that the algorithm can be \emph{restarted} several times---typically, this happens when more 
than one solution is requested or if restart strategy is activated by command-line options. (For simplicity, we did not include the code
for restarting the solver in Alg.~\ref{alg:yasmin}.)
In such cases, \InitialPropagation{} also handles  unit nogoods that have been learned in the previous execution.
The kernel invocation in line~2 specifies a grid of~{\tt b} blocks each composed of~{\tt t} threads.
The mapping is one-to-one between threads and unitary nogood.
In particular, if $k$ is the number of  unitary nogoods,
{\tt b}=$\lceil k/TPB \rceil$ and {\tt t}=$TPB$, where $TPB$ is the number of threads-per-block specified via command-line option.
The loop in lines~4--14 computes the answer set, if any.
Propagation is performed by the procedure \PropagateAndCheck{} in line~5, which also checks whether nogood violations occur. 
To better exploit the SIMT parallelism and maximize the number of concurrently active threads,
in each device computation the workload has to
be divided among the threads of the grid as  uniformly as possible.
To this aim, \PropagateAndCheck{} launches multiple kernels: one kernel deals with all nogoods with exactly two literals;
a second one processes the nogoods composed of three literals, and a further kernel processes all remaining nogoods.  
In this manner, threads of the same grid process a uniform number of atoms, reducing the divergence between them
and minimizing the number of inactive threads. Moreover, because, as mentioned, nogoods of the same length are stored contiguously,
threads of the same grid  are expected to realize coalesced accesses to global memory.
A more detailed description of the third of such device functions is given in Sect.~\ref{sect:PropAndCheck3}.
A similar technique is used in \PropagateAndCheck{} to process those nogoods that are learned at run-time
through the conflict analysis step (cf.~Sect.~\ref{sect:Learning}).
These nogoods are partitioned depending on their cardinality and processed by different kernels, accordingly.
In general, if $n$ is the number of nogoods of one partition, the corresponding kernel
has  {\tt b}=$\lceil n/TPB \rceil$ blocks of {\tt t}=$TPB$ threads each. Each thread processes one learned 
nogood. 

Propagation stops because either a fixpoint is reached (no more propagations are possible) or one or more conflicts occur.
In the latter case, if the current decision level is the top one the solver ends: no solution exists (line~6).
Otherwise, (lines 7--9) conflict analysis (\Learning{}) is performed and then the solver backjumps to a previous decision point (line~9).
The learning procedure is describes in Sect.~\ref{sect:Learning}. 
A specific kernel \Backjump{} takes care of  updating the value of $cdl$  and the array that stores the assignment.
A mapping one-to-one between threads and atoms in~$A$ is used.

On the other hand, if no conflict occurs and $A$ is not complete, a new \Decision{} is made (line~11).
As mentioned, the purpose of this kernel is to determine an unassigned atom $p$  which is head of an applicable rule~$r$.
All candidates $p$ and applicable~$r$
are evaluated in parallel  according to a typical heuristics to rank the atoms.
Possible criteria, selectable by command-line options, 
use the number of positive/negative occurrences
of atoms in the program (by either simply counting the occurrences or
by applying the Jeroslow-Wang heuristics) or the ``activity'' of atoms~\cite{SAThandbook}.
The first access to global memory to retrieve needed data is done in coalesced manner (a mapping one-to-one between threads and rules is used).
Then, a logarithmic parallel reduction scheme, implemented using thread-shuffling to avoid further accesses to global memory,
yields the rule $r$ with highest ranking.
Its head is selected and set true in the assignment.
\Decision{} also communicates to the solver whether no applicable rule exists (line~12).
In this case all unassigned heads in~$\Pi$ are assigned false (by the kernel \CompleteAssignment{} in line~13).
A successive invocation of \PropagateAndCheck{} validates the answer set and the solver ends in line~14.

\subsection{The propagate-and-check procedure}\label{sect:PropAndCheck3}
After each assignment of an atom of the current partial assignment $A$, each nogood $\delta$ needs to be analyzed to detect
whether: {(1)} it is violated, or
{(2)} there is exactly one literal $p$ in it that is  unassigned in~$A$,
in which case an inference step adds $\overline{p}$ to~$A$ (cf., Sect.~\ref{sect:CDASP}). 
The procedure is repeated until a fixpoint is reached. As seen earlier, this task is performed by the kernels launched by the 
procedure \PropagateAndCheck{}.

Alg.~\ref{alg:propandcheck} shows the device code of the generic kernel dealing with nogoods of length greater than three
(the others are simpler).
The execution of each iteration is driven by the atoms that have been  
assigned a truth value in the previous iteration (array $Last$ in Alg.~\ref{alg:propandcheck}).
Thus, each kernels involves a number of blocks that is equal to the number of such assigned atoms.
The threads in each block process the nogoods that share the same assigned atom. 
The number of threads of each  block is established by considering the
number of occurrences of each assigned atom in the input nogoods.
Observe
that the dimension of the grid may change between two consecutive invocations of the same kernel,
and, as such, it  is evaluated each time.
Specific data structures (initialized once during a pre-processing phase and stored in the
sparse matrix $\mathit{Map[][]}$ in Alg.~\ref{alg:propandcheck}) are used in order to
determine, after each iteration and for each assigned atom, 
which are the input nogoods to be considered.
A further technique is adopted to improve performance. Namely,
the processing of nogoods is realized by
implementing a standard technique based on \emph{watched literals}~\cite{SAThandbook}.
In this case, each thread accesses the watched literals of a nogood and acts accordingly.
The combination of nogood sorting and the use of watched literals,  improves the workload balancing among threads
and mitigates thread divergence.
(Watched literals are exploited also for long learned nogoods.)

Concerning Alg.~\ref{alg:propandcheck}, each thread of the grid first retrieves one of the atoms propagated
during the previous step (line~1). Threads of the same block obtain the same atom~$L$.
In line~2, threads accesses the data structure $Maps$, mentioned  earlier,
to retrieve the number $ngInBlock$ of nogoods to be processed by the block.
In line~5 each thread of the block determines which nogood has to be processed and retrieves its watched literals (lines~6-7).
In case one or both literals belongs to the current assignment~$A$, suitable substitutes are sought for (lines~10 and~14). 
Violation might be detected (lines~12 and~19, resp.) or propagation might occur (lines~16--18).
Notice that, concurrent threads might try to propagate the same atom (possibly with different sign), originating race conditions.
The use of atomic functions (line~16) allows one nondeterministically chosen thread $t$ to perform the propagation. Other threads may discover
agreement or detect inconsistency w.r.t.\ the value set by~$t$ (line~19).
In line~17 the thread~$t$ updates the set $Next$ of propagated atoms (to be used in the subsequent iteration) and stores (line~18)
information needed in future conflict analysis steps (by means of \mkdlbitmap{}, to be described in Sect.~\ref{sect:Learning})
and concerning the causes of the propagation. 

{\begin{algorithm}[tb]
\DontPrintSemicolon
\CommentSty{\color{blue}}
	\caption{\label{alg:propandcheck}Device code implementing propagation and nogood check ~~~\hfill(simplified)}
	{\bf procedure} {\sc nogood\_check}($\mathit{Last}$, $\mathit{Next}$ : ArrayOfLits, $\mathit{Map}$ : AtomsNogoodsMatrix, $A$: Assignment) \\
\nl	$L \gets \mathit{Last}[blockIdx.x]$\tcc*[f]{each block processes one of the propagated lits}\\
\nl	$\mathit{ngInBlock} \gets ~|\mathit{Map}[L]|$\tcc*[f]{get the number of nogoods in which $L$ occurs}\\
\nl	$i \gets \mathit{threadIdx}.x$\tcc*[f]{each nogood in which $L$ occurs is treated by a thread}\\
\nl	\If{$i < \mathit{ngInBlock}$}{
\nl		$\delta \gets \mathit{Map}[L][i]$\tcc*[f]{get the nogood}\\
\nl		$w_1 \gets \mathit{watched}_1[\delta]$\tcc*[f]{copy the two watched lits in registers}\\
\nl		$w_2 \gets \mathit{watched}_2[\delta]$\\
	\nl		\lIf(\tcc*[f]{satisfied nogood, thread exits}){$\overline{w_1}\in A \vee \overline{w_2}\in A$}{{\bf return}}
\nl		\uIf{$w_1\in A \wedge w_2\in A$}{
\nl			\uIf{{\bf exists}  $w\in \delta$ such that $w\not\in A \wedge\overline{w}\not\in A$}{
\nl				$w_1 \gets w$
			}\nl \lElse(\tcc*[f]{nogood violation}){ $Viol \gets true$ }
	       	}
	\nl		\uIf(\tcc*[f]{the case $w_2\not\in A\wedge w_1\in A$ is analogous (omitted)}){$w_1\not\in A\wedge w_2\in A$}{
\nl			\uIf{{\bf exists}  $u\in \delta$ such that $w_1\not=u \wedge u\not\in A \wedge \overline{u}\not\in A$}{
\nl				$w_2 \gets u$
			} \uElse(\tcc*[f]{first thread propagates (others may agree or cause violation)}){
\nl				\uIf(\tcc*[f]{returns true if found not disagreeing}){$atomicSet(A, w_1, cdl)$}{ 
\nl					$\mathit{Next} \gets \mathit{Next} \cup \{w_1\} $\tcc*[f]{update set of propagated lits}\\
\nl					$\mathit{Deps}[w_1] = \mkdlbitmap(w_1,\delta)$\tcc*[f]{set dependencies of $w_1$}\\
   				}\nl \lElse(\tcc*[f]{if disagreeing, it is a violation}){ $Viol \gets true$ }
			}
		}
\nl $\mathit{watched}_1[\delta] \gets w_1$
~;~~ $\mathit{watched}_2[\delta] \gets w_2$\tcc*[f]{update the two nogoods}\\
} 
\medskip

{\tt\_\_inline\_\_} ~~{\bf procedure} {\sc mk\_dl\_bitmap}($w$ : Literal,  $\delta$: Nogood) \\
\nl $\reset(res)$\tcc*[f]{empty set = null bitmap}\\
\nl \myForEach(\tcc*[f]{collect causes of propagation of $w\in\delta$}){$x\in \delta\setminus\{w,\overline{w}\}$}{
\nl \lIf(\tcc*[f]{if $dl(x)>1$, $x$ is not an input unit}){$dl(x)>1$}{
 $res \gets res ~|~ \mathit{Deps}[x]$
}
}       
\nl {\bf return} $res$
\end{algorithm}
}

{\begin{algorithm}[tb]
\DontPrintSemicolon
\CommentSty{\color{blue}}
	\caption{\label{alg:res-learning}Resolution based learning schema in {\sc clasp}~\cite{Gebser2012Practice}}
{\bf procedure} {\sc res-learning}({$\delta$: Nogood, $\Delta$: SetOfNogoods, $A$: Assignment})\\
\nl\myWhile{\mbox{{\bf exists} } $\sigma\in\delta$ \mbox{{\bf such that}} $\delta\setminus A = \{\sigma\}$}
{ 
{\tcc{get the decision level $\kappa$  of the last but one assigned literal in $\delta$}
	\nl $\kappa \gets \max(\{dl(\rho)\mid\rho\in\delta\setminus\{\sigma\}\}\cup\{0\})$\\
\nl\uIf(\tcc*[f]{there is another lit in $\delta$ decided at level $dl(\sigma)$}){$\kappa=dl(\sigma)$}
  {
\nl\uLet{$\varepsilon\in\Delta$ \mbox{{\bf such that}} $\varepsilon\setminus A=\{\overline{\sigma}\} \mbox{\,{\bf in}}$}
    {
\nl $\delta\gets (\delta\setminus\{\sigma\})\cup(\varepsilon\setminus\{\overline{\sigma}\})$
\tcc*[f]{resolution step between $\delta$ and $\varepsilon$}
    }
  }
\nl\lElse{{\bf return}$(\delta,\kappa)$}
}
}
\end{algorithm}
}

{\begin{algorithm}[tb]
\DontPrintSemicolon
\CommentSty{\color{blue}}
	\caption{CUDA device code using warp-shuffling for fwd-learning\label{alg:shufflefwdlearning} ~~~\hfill(simplified)}
{
{\bf procedure} {\sc fwd-learning}($\delta$ : Nogood, $\Delta$ : SetOfNogoods, $A$ : Assignment, $\dependencies$ : ArrayOfBitmaps)\\
	\nl	$i \gets \mathit{threadIdx}.x$\tcc*[f]{id of the thread (for simplicity in D1 grid)}\\
	\nl  {\tt\_\_shared\_\_} ~~ $\mathit{sh\_bitmap}[\mathit{warpSize}]$\tcc*[f]{array of bitmaps shared among threads}\\
	\nl  $lane \gets i ~\%~ \mathit{warpSize}$\tcc*[f]{lane of the thread in its warp}\\
	\nl  $wid \gets i / \mathit{warpSize}$\tcc*[f]{id of thread's warp}\\
	\nl	\reset($\mathit{vbitmap}$)\tcc*[f]{each thread resets its private bitmap}\\
\nl	\syncthreads()\tcc*[f]{synchronization barrier}\\
\nl\myWhile(\tcc*[f]{collect dependencies of all atoms in $\delta$ from global memory}){$i < \mid\delta\mid$}
{
	\nl $\mathit{atom} \gets \mbox{ the $i$-th atom in } \delta$\tcc*[f]{strided access to memory}\\
	\nl	$\mathit{vbitmap} \gets \mathit{vbitmap} ~|~ \dependencies[\mathit{atom}]$\tcc*[f]{bit-a-bit disjunction: collects deps}\\
	\nl	$i \gets i + \mathit{blockDim.x}$\tcc*[f]{next stride}\\
}
\nl	\syncthreads()\\ 
   	\tcc{logarithmic reduction using shuffling within each warp:}
	\nl\myFor{($\mathit{offset} \gets \mathit{warpSize}/2$; $\mathit{offset}>0$; $\mathit{offset}/=2$)}{
	\nl	$\mathit{vbitmap} \gets \mathit{vbitmap} ~|~ \mbox{\tt\_\_shfl\_down\_sync}(\mbox{\tt 0xFFFFFFFF}, \mathit{vbitmap}, \mathit{offset})$
  }
\nl\lIf{lane=0}{
	$\mathit{sh\_bitmap}[wid] \gets \mathit{vbitmap}$\tcc*[f]{store reduced value in shared memory}
}
\nl	\syncthreads()\tcc*[f]{wait for all partial reductions}\\
\tcc{read from shared memory only if that warp participates:}
	\nl $\mathit{vbitmap} \gets (\mathit{threadIdx}.x < \mathit{blockDim.x} / \mathit{warpSize}) \mbox{\,?\,} \mathit{sh\_bitmap}[lane] : 0$\\
	\nl\uIf(\tcc*[f]{the first warp performs the final reduction}){wid=0}{
\nl\myFor{($\mathit{offset}\gets \mathit{warpSize}/2$; $\mathit{offset}>0$; $\mathit{offset}/=2$)}{
\nl $\mathit{vbitmap} \gets \mathit{vbitmap} ~|~ \mbox{\tt\_\_shfl\_down\_sync}(\mbox{\tt 0xFFFFFFFF}, \mathit{vbitmap}, \mathit{offset})$
}
}
\nl	\syncthreads()\\
	\tcc{Here $\mathit{vbitmap}$ encodes all dependencies of all literals in $\delta\setminus\{\sigma\}$}
	\nl\uIf(\tcc*[f]{add deps for $\overline{\sigma}$ in $\varepsilon$  (gathered during propagation)}){$\mathit{threadIdx}.x = 0$}{ 
	\nl	$\mathit{sh\_bitmap}[0] \gets \mathit{vbitmap} ~|~ \dependencies[\overline{\sigma}]$\\
	\nl	$\mathit{backjump\_dl} \gets$ \leftmostsetbit($\mathit{sh\_bitmap}[0]$)\tcc*[f]{gets the level to backjump to}\\
}
\nl	\syncthreads()\\
\tcc{Store the learned nogood: the threads of the first warp store in global memory, in coalesced way, the relevant decision literals}
	\nl\uIf{$\mathit{threadIdx}.x < \mathit{warpSize}$}{  
	\nl $i \gets \mathit{threadIdx}.x$\\
	\nl\myWhile(\tcc*[f]{for all decision levels}){$i<\mathit{max\_dl}$}{
		\nl\uIf(\tcc*[f]{if conflict depends on the $i$-th decision}){$(\mathit{sh\_bitmap}[0] ~\&~ 2^{i+1})$}{
\nl	$\mathit{new\_nogood} \gets \mathit{new\_nogood} \cup \getithdecision(i)$
}
	\nl $i \gets i + \mathit{warpSize}$\\
}
}
}
\end{algorithm}
}

\noindent{\scriptsize
\begin{table}[tb]
{\centerline{\scriptsize
\begin{tabular}{ccc}
\begin{tabular}{|r|l|r|r|}
\hline
{ID} &{Instance} & {Nogoods}  & {Atoms}     \\
\hline
I0 & 0001-visitall      & 42286 &    17251       \\
I1 & 0003-visitall      & 40014 &    16337       \\
I2 & 0167-sokoban      & 68585 &    29847       \\
I3 & 0010-graphcol       & 37490 &   15759        \\
I4 & 0007-graphcol       & 37815 &   15889        \\
I5 & 0589-sokoban      & 76847 &     33417      \\
I6 & 0482-sokoban      & 84421 &    36639       \\
I7 & 0345-sokoban      & 119790 &    51959       \\
I8 & 0058-labyrinth      & 228881 &    84877       \\
I9 & 0039-labyrinth      & 228202 &    84633       \\
I10 & 0009-labyrinth      & 228859 &   84865        \\
\hline
\end{tabular}
&
\begin{tabular}{|r|l|r|r|}
\hline
{ID} &{Instance} & {Nogoods}  & {Atoms}     \\
\hline
I11 & 0023-labyrinth      & 228341 &   84677        \\
I12 & 0008-labyrinth      & 229788 &   85189        \\
I13 & 0041-labyrinth      & 228807 &   84853        \\
I14 & 0007-labyrinth      & 229539 &   85100        \\
I15 & 0128-ppm       & 589884 &      14388     \\
I16 & 0072-ppm      & 591542   &     14679      \\
I17 & 0153-ppm      & 721971 &       16182    \\
I18 & 0001-stablemarriage     & 975973 &   63454        \\
I19 & 0005-stablemarriage     & 975945 &   63441        \\
I20 & 0010-stablemarriage     & 975880 &   63415        \\
I21 & 0004-stablemarriage     & 975963 &   63453        \\
\hline
\end{tabular}
&
\begin{tabular}{|r|l|r|r|}
\hline
{ID} &{Instance} & {Nogoods}  & {Atoms}     \\
\hline
I22 & 0003-stablemarriage     & 975930 &   63438        \\
I23 & 0009-stablemarriage     & 975954 &   63447        \\
I24 & 0002-stablemarriage     & 975907 &   63430        \\
I25 & 0006-stablemarriage     & 975953 &   63446        \\
I26 & 0008-stablemarriage     & 975934 &   63439        \\
I27 & 0007-stablemarriage     & 976047 &   63486        \\
I28 & 0061-ppm      & 1577625 &       24465    \\
I29 & 0130-ppm      & 1569609 &       24273    \\
I30 & 0121-ppm       & 2208048 &      28776     \\
I31 & 0129-ppm       & 4854372 &      43164     \\
	&&&\\
\hline
\end{tabular}
\end{tabular}
	}\caption{Some instances used in experiments.
	The table shows: shorthand IDs, instance names (taken from~\cite{DFPV16}), the numbers of nogoods/atoms given as input to the solving phase of {\sc yasmin}.}\label{tab:instances}}
\end{table}
}

\subsection{The learning procedure}\label{sect:Learning}

As mentioned, the \Learning{} procedure is used to resolve a conflict detected by \PropagateAndCheck{} and to
identify a decision level
the computation should backjump to, in order to remove the violation.
 The analysis usually performed in ASP solvers such as {\sc clingo}~\cite{Gebser2012Practice}
 demonstrated rather unsuitable to SIMT parallelism.
 This is  due to the fact that a sequential sequence of  resolution-like steps must be encoded.

 In the case of the parallel solver {\sc yasmin}, more than one conflict might be detected by \PropagateAndCheck{}.
The solver selects one or more of them 
(heuristics can be applied to perform such a selection, for instance, priority can be assigned to shorter nogoods.)
For each selected conflict, a grid of a single block, to facilitate synchronization, is run to perform 
 a sequence of resolution steps, starting from the conflicting nogood (say, $\delta$),
 and proceeding backward, by resolving upon the last but one assigned atom~$\sigma\in\delta$.
 The step involves $\delta$ and a nogood $\varepsilon$ including~$\overline{\sigma}$.
 Resolution steps end as soon as the last two assigned atoms in $\delta$ correspond to different decision levels.
 This approach identifies the \emph{first~UIP}~\cite{SAThandbook}.
 Alg.~\ref{alg:res-learning} shows the pseudo-code of such procedure (see also~\cite{SAThandbook,Gebser2012Practice} for the technical details).
 The block contains a fixed number (e.g., 1024) of threads and every thread takes care of
 one atom (if there are more atoms than threads involved in the learning, atoms are equally partitioned among threads).
For each analyzed conflict,  a new nogood is learned and added to~$\Delta$.
In case of multiple learned nogoods involving different ``target'' decision levels, the lowest level is selected.

In order to remove the computational bottleneck represented by this kind of learning strategy we designed an alternative,
parallelizable, technique.
The basic idea consists in collecting, during the propagation phase, information useful to speed up conflict analysis,
affecting as little as possible, performance of propagation.
A bitmap $\dependencies[p]$ is associated to each atom $p$. The $i$-th bit of $\dependencies[p]$ is set 1 if the assignment of $p$
depends (either directly or transitively) on the atom decided at level~$i$.
Hence, when an atom $q$ is decided at level $j$, $\dependencies[q]$ is assigned the value $2^{j-1}$ (by
the procedure \Decision{}).
Whenever propagation of an atom $w_1$ occurs (see Alg.~\ref{alg:propandcheck}, line 18) the function \mkdlbitmap{} computes 
the bit-a-bit disjunction of all bitmaps associated to all other atoms in~$\delta$.
To maximize efficiency this computation is performed by a group of threads, exploiting shuffling, through a 
logarithmic parallel reduction scheme.
Alg.~\ref{alg:shufflefwdlearning} shows the code of the new learning procedure.
The kernel \fwdlearning{} is run by a grid of a single block, where each thread processes an atom of the conflicting nogood~$\delta$.
Initially, each thread determines the index of its warp (line~4) and its relative position in the warp (line~3).
After a synchronization barrier (line~6) each thread retrieves the bitmaps of one or more atoms of~$\delta$.
The disjunction of these bitmaps is stored in the private variable $\mathit{vbitmap}$.
Then, each warp executes a logarithmic reduction scheme (lines 12--14)
to compute a partial result in shared memory (allocated in line~2).
At this point, the first warp performs a last logarithmic reduction (lines 17--19) combining all partial results.
After a  synchronization barrier, thread~0 adds the dependencies relative to $\overline{\sigma}$ in $\varepsilon$ (line~22)
and determines the decision level to backjump to (line~23).
Finally, the learned nogood in built up using the bitmap $\mathit{sh\_bitmap}[0]$ and stored in global memory in coalesced 
way (lines 25--30).

{\footnotesize
\begin{figure}[tb]
{\centerline{\includegraphics[width=0.80\linewidth]{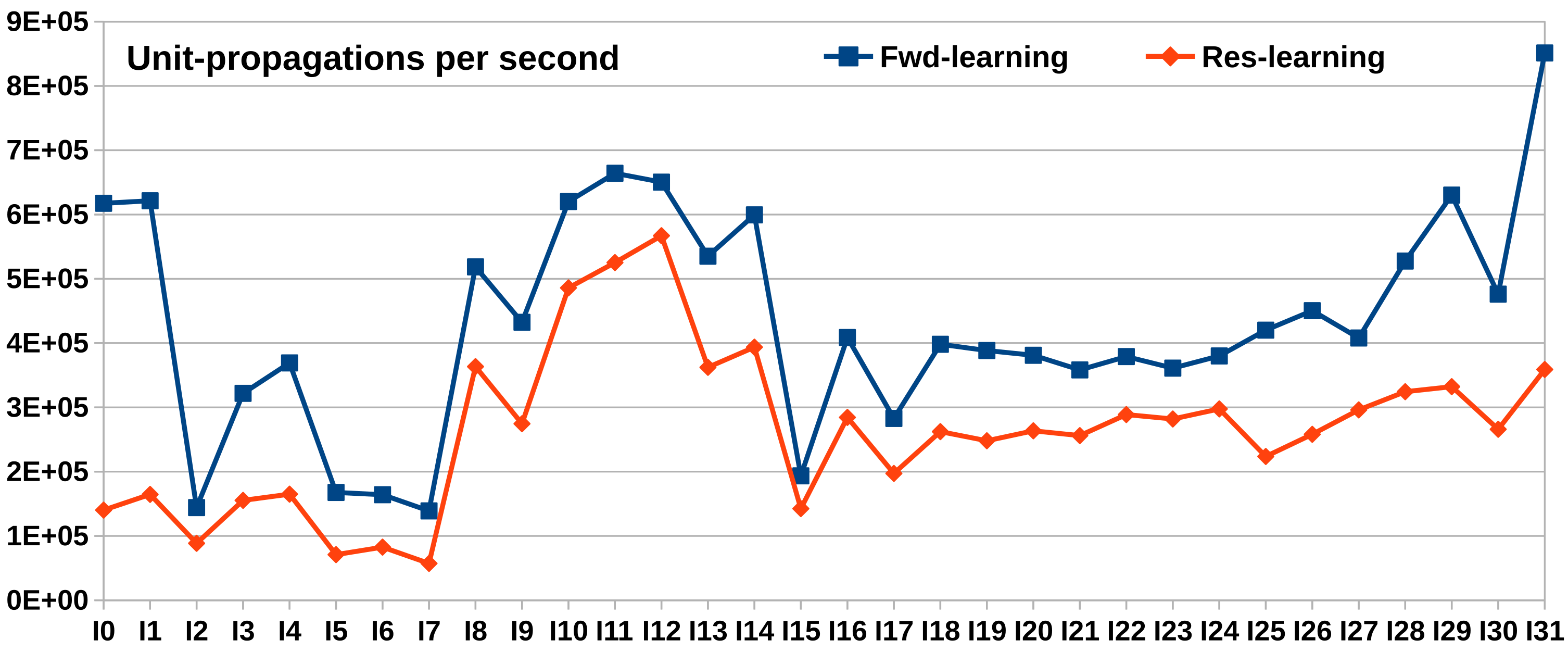}}
\vspace*{1.0ex}}
{\centerline{\includegraphics[width=0.80\linewidth]{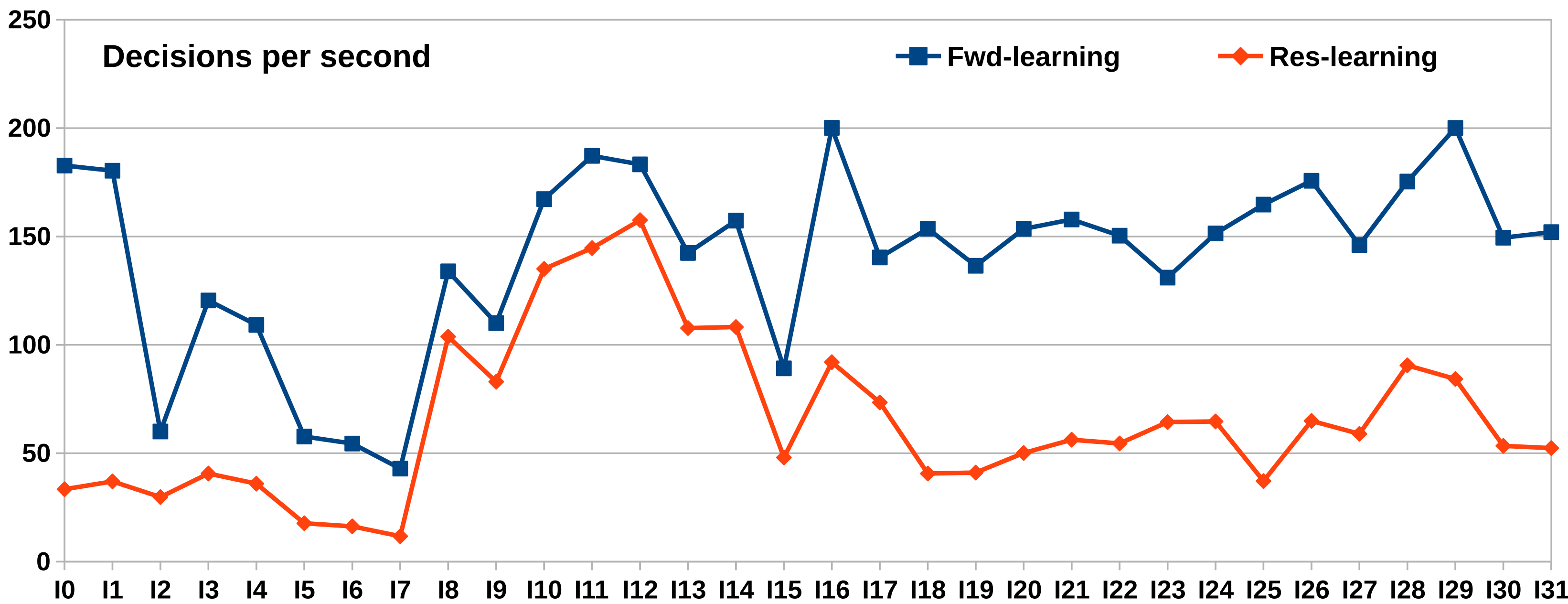}}
\caption{\label{fig:upPS}Performance of the two versions of {\sc yasmin} (using Res-learning and Fwd-learning). Number of propagations per second~(top) and number of decisions per second~(bottom).}}
\end{figure}
}

{\footnotesize
\begin{figure}[tb]
{\centerline{\includegraphics[width=0.80\linewidth]{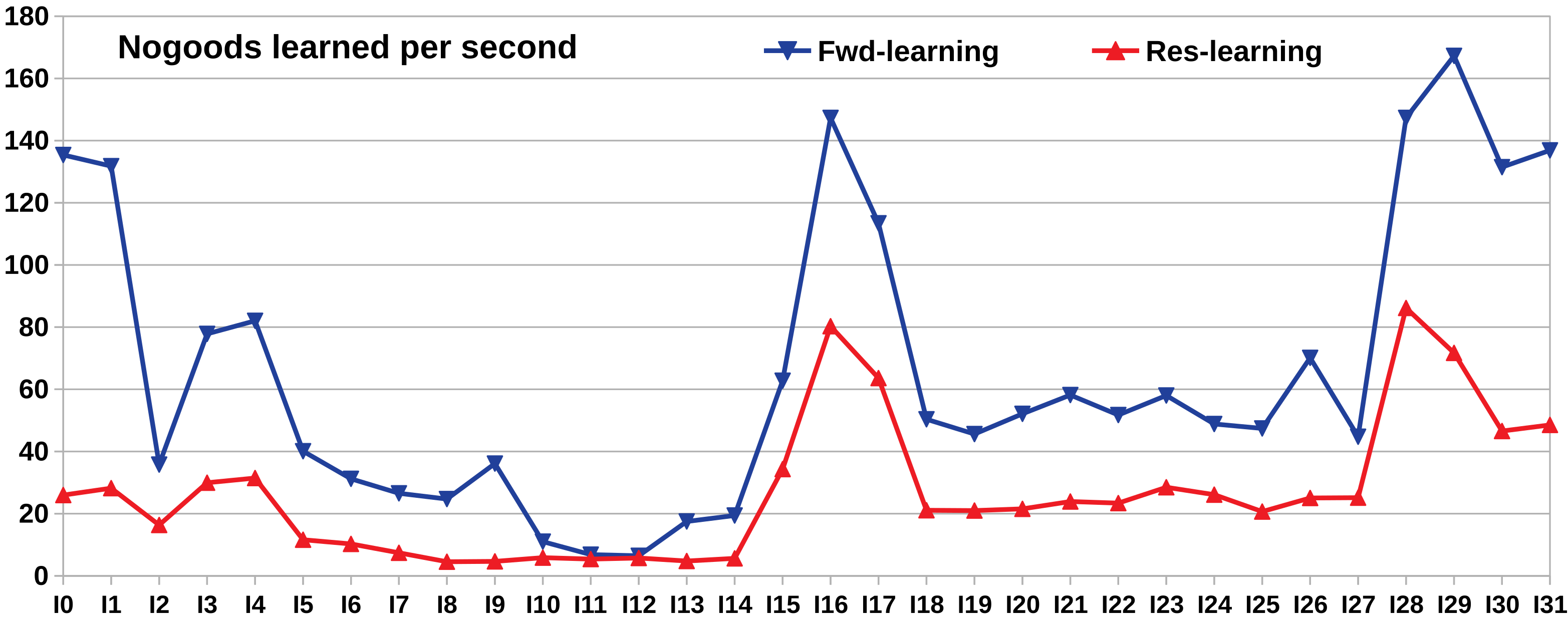}}
\vspace*{1.0ex}}
{\centerline{\includegraphics[width=0.80\linewidth]{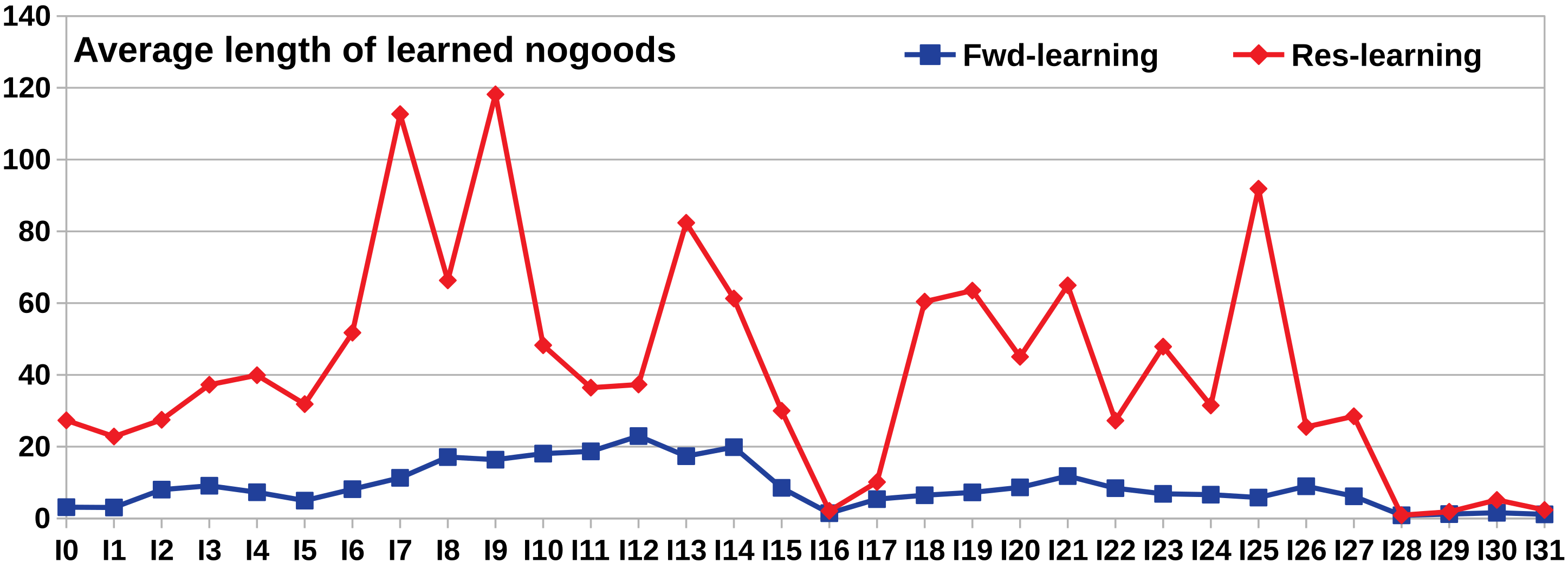}}
\caption{\label{fig:LearnedAndAvgLen}Learned nogoods using Res-learning and Fwd-learning. Number of learned nogoods per second~(top) and their average length~(bottom).}}
\end{figure}
}

\section{Experimental Results}\label{sect:runs}
In this section we briefly report on some experiments we run to compare the two learning techniques described in the previous section.
Table~\ref{tab:instances} shows a selection of the instances (taken from~\cite{DFPV16}) we used.
For each instance the table indicates, together with
an ID, the number of nogoods and the number of atoms.

Experiments were run on a Linux PC (running Ubuntu Linux v.19.04), used as host machine, and using
as device a Tesla K40c Nvidia GPU with these characteristics:
2880 CUDA cores at 0.75 GHz, 12GB of global device memory.
We used on such GPU the CUDA runtime version 10.1. The compute capability was~3.5.

Fig.~\ref{fig:upPS} compares the two versions of {\sc yasmin} solver,
differing only on the used learning procedure. Comparison is made w.r.t.\ the number of 
propagations per second and the number of decisions per second performed by the solver.
The new learning strategy outperforms the resolution-based one on all instances.
The plots in Fig.~\ref{fig:LearnedAndAvgLen} compare the performance of the two learning procedure in terms of
their outcomes. Also from this perspective \fwdlearning{} exhibits better behavior, 
producing smaller nogoods in shorter time.
Notice that results of the same kind have been obtained with different selection heuristics and varying the parameters of kernel 
configuration (e.g., number of threads-per-block, grid and block dimensions, etc.). 
Moreover, results of experiment run on different GPUs are in line with those reported.

\section*{Conclusions}
This paper we described the main traits of a CUDA-based solver for Answer Set Programming.
The fact that the algorithms involved in ASP-solving present
an irregular and low-arithmetic intensity, usually  combined with  data-dependent control flows, 
makes it difficult to achieve high performance without adopting proper sophisticated solutions and 
fulfilling suitable programming directives. 
In this paper we dealt with the basic software  architecture  of a parallel prototypical solver 
with the main aim of demonstrating that GPU-computing can be exploited in ASP solving.
Much is left to do in order to obtain a full-blown parallel solver able to compete with the  state-of-the-art existing solvers.
First, effort have to be made in enhancing the parallel solver with the collection of heuristics proficiently used
to guide the search in sequential solvers. Indeed, experimental comparison~\cite{DFPV16} show that good heuristics
might be the most effective component of a solver.
Second, the applicability of further techniques and refinements have to be investigated. For instance,
techniques such as \emph{parallel lookahead}~\cite{DFP18}, \emph{multiple learning}~\cite{FVictcs14}, should be considered.
Also the possibility of developing a distributed parallel solver that operates on multiple GPUs represents a challenging theme of research.

\end{document}